\documentclass{elsarticle}

\journal{Pattern Recognition}
\usepackage{url}
\usepackage{subcaption}
\usepackage{amssymb}
\usepackage{amstext}
\usepackage{amsmath}
\usepackage{amsthm}
\usepackage[usenames,dvipsnames]{xcolor}
\usepackage{multirow}
\usepackage{dcolumn}
\usepackage[english]{babel}
\usepackage[utf8]{inputenc}
\usepackage{graphicx}
\graphicspath{{./figs/}}
\DeclareGraphicsExtensions{.pdf,.png}
\usepackage{algorithm}
\usepackage{algorithmic}
\usepackage{verbatim}
\usepackage[colorinlistoftodos,textwidth=4cm]{todonotes}
\usepackage[normalem]{ulem}

\usepackage{ctable}
\usepackage{setspace}

\usepackage{soul}

\DeclareMathOperator{\CAE}{SAE}

\begin{document}


\newcommand{\review}[1]{\textcolor{black}{#1}}


\begin{frontmatter}

\title{A selectional auto-encoder approach for document image binarization}

\author[add1,add2]{Jorge Calvo-Zaragoza\corref{cor1}}
\ead{jcalvo@prhlt.upv.es}
\cortext[cor1]{Corresponding author: 
Tel.: +346-66-837771}

\author[add3]{Antonio-Javier Gallego}
\ead{jgallego@dlsi.ua.es}

\address[add1]{Schulich School of Music, McGill University, 555 Sherbrooke St W, Montreal, QC H3A 1E3, Canada}
\address[add2]{PRHLT Research Center, Universitat Polit\`ecnica de Val\`encia, Valencia, 46022, Spain}

\address[add3]{Department of Software and Computing Systems, University of Alicante, Carretera San Vicente del Raspeig s/n, 03690 Alicante, Spain}

\date{\today}

\begin{abstract}
Binarization plays a key role in the automatic information retrieval from document images. This process is usually performed in the first stages of documents analysis systems, and serves as a basis for subsequent steps. Hence it has to be robust in order to allow the full analysis workflow to be successful. Several methods for document image binarization have been proposed so far, most of which are based on hand-crafted image processing strategies. Recently, Convolutional Neural Networks have shown an amazing performance in many disparate duties related to computer vision. In this paper we discuss the use of convolutional auto-encoders devoted to learning an end-to-end map from an input image to its selectional output, in which activations indicate the likelihood of pixels to be either foreground or background. Once trained, documents can therefore be binarized by parsing them through the model and applying a global threshold. This approach has proven to outperform existing binarization strategies in a number of document types.
\end{abstract}


\end{frontmatter}

\section{Introduction}
\label{sec:introduction}
Image binarization consists in assigning a binary value to every single pixel of an image. Within the context of document analysis systems, the main objective is to distinguish the \emph{foreground} (meaningful information) from the \emph{background}.

Binarization plays a key role in the workflow of many document analysis and recognition systems \cite{LOULOUDIS20083758,saabni2014text,HE20154036,calvo2015avoiding,GIOTIS2017310}. It not only helps to reduce the complexity of the task but is also advisable for procedures involving morphological operations, detection of connected components, or histogram analysis, among others. Many methods have been proposed to accomplish this task. However, it is often complex to attain good results because documents may contain several difficulties---such as irregular leveling, blots, bleed-through, and so on---that may cause the process to fail.

In addition to all these obstacles, it is convenient to emphasize that it is very difficult for the same method to work successfully in a number of document styles, since the set of potential domains is very heterogeneous. In order to deal with this situation, we discuss a framework with which to binarize image documents based on machine learning. That is, a ground-truth of examples is used to train a model to perform the binarization task. This allows using the same approach in a wide range of documents, provided there is specific ground-truth data to train a new model for each document type.

Specifically, we make use of Convolutional Neural Networks (CNN) \cite{zeiler2014visualizing}. These networks involve multi-layer architectures that perform a series of transformations (convolutions) to the input signal. The parameters of these transformations are adjusted through a training process. CNN have dramatically improved the state of the art in many tasks such as image, video, and speech processing~\cite{lecun15deep}. Thus, its use for document binarization is promising. In this case, we consider an image-to-image convolutional architecture, which is trained to transform an input image into its binarized version. 

Our experiments focus on testing this strategy in different document types, namely Latin text documents, palm leaf scripts, Persian documents, and music scores. We also compare the approach against other classical and state-of-the-art algorithms for binarization, showing that this approach leads to a significant improvement.

The remainder of the paper is structured as follows: related works to document image binarization are introduced in Section~\ref{sec:background}; the image-to-image binarization framework based on convolutional models is described in Section~\ref{sec:method}; the experiments are presented in Section~\ref{sec:experiments}, including model configuration, training strategies, comparison with existing techniques, and cross-document adaptation; and finally, the current work concludes in Section~\ref{sec:conclusions}.

\section{Background} \label{sec:background}
The most straightforward procedure for image binarization is to resort to simple thresholding, in which all pixels under a certain grayscale value are set to $0$, and those above to $1$. This threshold can be fixed by hand, yet algorithms such as Otsu's \cite{otsu1975threshold} automatically estimate a value according to the input image. However, as the complexity of the document to process increases, this simple procedure usually leads to poor or irregular binarization, and so it is preferable to resort to other kind of approaches.

Instead of computing a single global threshold, different threshold values might be obtained for every pixel by taking into account the features of its local neighborhood (defined by a window centered around the pixel). Niblack's algorithm~\cite{niblack1985introduction} was one of the first binarization processes following this approach. Let $m$ and $s$ be the mean and standard deviation, respectively, of the considered neighboring region, it computes the threshold as $T = m + k \cdot s$, where $k$ is a parameter to tune. 

Most common extensions to this approach are Sauvola's~\cite{sauvola2000adaptive} and that proposed by Wolf et al.~\cite{wolf02text}, which try to boost the accuracy of the binarization by considering more complex equations for the adaptive threshold. The former assumes that foreground pixels are closer to black than background pixels, whereas the latter normalizes contrast and mean gray-level of the considered neighborhood. Recently, Lazzara and Geraud proposed an efficient multi-scale version of Sauvola's method (Sauvola MS) to deal better with documents depicting components of different sizes \cite{lazzara2014efficient}. 

Posterior to these methods, several procedures for document image binarization have been proposed, which  add more steps to the pipeline. Gatos et al. method~\cite{gatos06adaptive} is an adaptive procedure that follows several steps, namely a low-pass Wiener filter, estimation of foreground and background regions, and a thresholding. It ultimately applies a post-processing step to improve the quality of foreground regions and preserve stroke connectivity. The method proposed by Su et al.~\cite{su2013robust} uses an adaptive image contrast as a combination of the local image contrast and the local image gradient, which makes it more robust against document degradations. The contrast map is then combined with an edge detector to identify the boundaries of foreground elements. Then, the document is finally binarized by using a local threshold based on the values of those boundaries. Howe's method \cite{howe2011laplacian} considers the Laplacian operator to compute the local likelihood of foreground and background labels, and Canny algorithm to detect discontinuities. It then formulates the binarization task as a global energy minimization problem. The method contains several parameters, which can be automatically tuned following a subsequent work \cite{howe2013document}. Very recently, Howe's method has been combined with a new preprocessing that boosts its performance. It linearly transforms the image onto a spherical surface in which concavities correspond to foreground in the original image. The concavities are estimated with the Hidden Point Removal operator \cite{katz2007direct}, which produces a probability for each pixel of the image to belong to a concavity. This stochastic representation is then used as input for Howe's binarization. This strategy was proposed by Kliger and Tal in the Document Image Binarization Contest 2016~\cite{PratikakisZBG16}, 
in which it was ranked first.

All aforementioned methods have been successful in their contexts. However, it is difficult for these strategies to generalize adequately to any type of document. It is, therefore, interesting to resort to binarization procedures that can be learned, in order to apply such strategies over the largest possible set of document domains as long as training data is available. That is, our idea is to consider algorithms that are able to binarize documents provided they have been trained for it.

In this regard, supervised learning techniques have been also studied for binarization tasks. The classification-based approach typically consists in querying every single pixel of the image, performing a feature extraction out of it and using a supervised learning algorithm to output a hypothesis about the two possible categories. Traditionally, the common choice has been to resort to Multi-Layer Perceptron~\cite{chi2001two,Hidalgo2005,kefali2014foreground}. Within this paradigm, it is obvious the consideration of deep neural networks---especially Convolutional Neural Networks (CNN) given their striking impact on the machine learning and computer vision communities~\cite{lecun15deep}. Some insights on the use of pixel-wise CNN for document image binarization can be found in the work of Pastor-Pellicer et al.~\cite{pastor-pellicer15insights}. The use of CNN has also proven to be robust when dealing with the binarization of document images that do not necessarily depict text information~\cite{calvo17binarization}.

Although these works have shown acceptable performance, this pixel-wise approach has two important drawbacks. The first is the high computational cost that requires labeling each pixel of the image, as it involves making as many predictions as pixels of the image. This is an important issue since binarization is usually a pre-processing step of a larger workflow, and such computational cost represents a bottleneck for the whole process. Secondly, each pixel is classified independently, without taking into account information about the labels assigned to their neighbors. That is, contextual information is somehow wasted. 

The pixel-wise approach, however, serves as an inspiration to go on considering learning-driven binarization methods. In this work we study a convolutional approach that tries to alleviate the aforementioned drawbacks. We propose the use of fully-convolutional auto-encoder models that are trained to learn a patch-wise mapping of the image to its corresponding binarization. Note that this does not consist in assigning the same label to all pixels of the input patch, but to perform a fine-grain categorization in which each pixel of the output gets a different activation value depending on whether the pixel must be labeled as background or foreground.  

Fully-convolutional models have been previously used for semantic segmentation \cite{ShelhamerLD17}, in which each pixel of an image is assigned a label corresponding to the category to which it belongs. Furthermore, auto-encoders have been considered for image denoising \cite{vincent2010stacked}, which is a task of a similar formulation than that of document binarization (color can be considered noise of the input image). Therefore, the use of these models seems promising for document image binarization.
Actually, fully-convolutional auto-encoders have been recently introduced for this purpose \cite{PengCN17}. In this work, we thoroughly evaluate the capabilities of these models for the document binarization task, studying several network topologies, using a number of datasets, performing in-depth analyses, and conducting experiments in cross-document scenarios.

\section{Selectional auto-encoder for document image binarization} \label{sec:method}
From a machine learning point of view, image binarization can be formulated as a two-class classification task at pixel level. The framework proposed follows this idea and, therefore, consists in learning to estimate which label must be given to every single pixel of an image. Since we are dealing with images of documents, we define the set of labels as \emph{foreground} and \emph{background}. As mentioned above, a way to implement this approach is to use a neural network to estimate which of these labels must be assigned to a pixel one by one.

In this paper we go a step further, considering an approach that uses a network topology that learns to perform an image-to-image processing. That is, for each input image, the model is devoted to binarizing it in just one step. This has a number of advantages such that, in this way, the classification of each pixel of the image is not computed independently, but also takes into account (indirectly) the label to be assigned to its neighbors. In addition, all the pixels within the input image are processed at the same time, thereby leading to higher efficiency.

This approach is inspired by classical adaptive algorithms. Nevertheless, these algorithms consider that the pixel value is static (i.e., its value in grayscale), whereas a different threshold is calculated depending on the local context. What the proposed framework does is to compute a different value for each pixel (neural activation) and fix a global threshold. Obviously, this activation is not obtained following a hand-crafted equation, which may serve well for a specific domain, but is learned through fully-convolutional auto-encoders.

Auto-encoders consist of feed-forward neural networks for which the input and output shape is exactly the same~\cite{hinton1994autoencoders}. The network is typically divided into two stages that learn the functions $f$ and $g$, which are called encoder and decoder functions, respectively. Traditionally, the network receives an input vector $x$ and it must minimize a divergence $L(x, g(f(x)))$. The hidden layers of the encoder perform a mapping of the input---usually decreasing its dimension---until an intermediate representation is attained. The same input is then subsequently recovered by means of the hidden layers of the decoder function. 

In their original formulation, auto-encoders were trained to learn the identity function, which might be useful as regards feature learning or dimensionality reduction because the encoder function provides a meaningful, compact representation of the input~\cite{Wang_2014}. In this work, however, we use an auto-encoder topology for a different purpose. Here the model is trained to perform a function such that $b : \mathbb{R}^{(w \times h)} \rightarrow [0,1]^{(w \times h)}$. In other words, it learns a selectional map over a $w \times h$ image that preserves the input shape. The selectional value (activation) of each pixel depends on whether the pixel belongs to the foreground or to the background. Given the nature of the output, we call this model \emph{Selectional Auto-Encoder} ($\CAE$). This selectional encoder-decoder approach has shown promising performance in related tasks such as contour detection \cite{YangPCL016,GALLEGO2017138} but its use for document binarization remains widely unexplored.

Note that the approach is fully convolutional in the sense that the prediction can be done through successive convolutions and sampling operations, without any fully-connected layer. In other words, the approach just applies successive transformations to the input image. Theoretically, this also implies that the input and output layers may be of an arbitrary size. In practice, however, this could require very high computational resources, and so here we assume that the network has pre-established input and output sizes.

The underlying transformations of the $\CAE$ are learned through a training process, rather than being pre-established, with the aim at binarizing the document image. 
The training stage consists of providing examples of images and their corresponding binarized ground-truth. Since an $\CAE$ is a type of feed-forward network, the training process can be carried out by conventional means \cite{glorot2010understanding}.

\begin{figure}[ht]
\centering
	\includegraphics[width=.9\columnwidth]{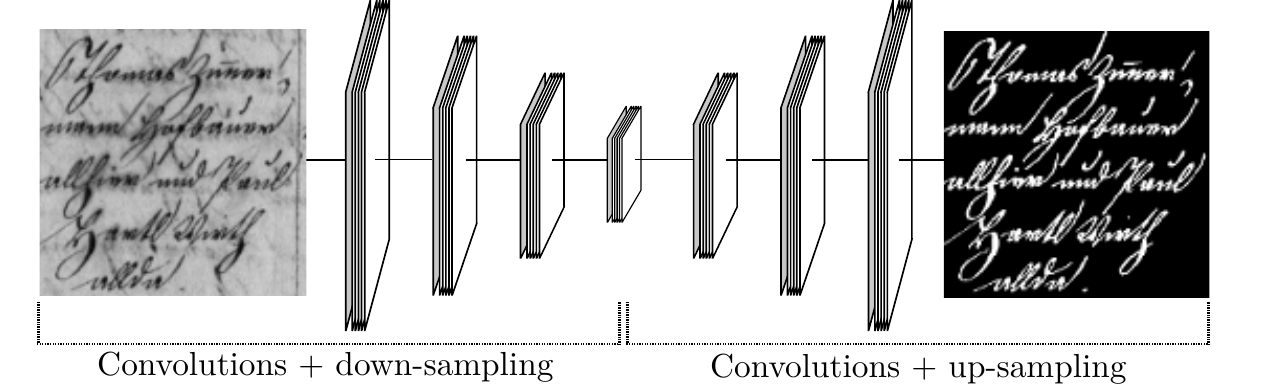}
	\caption{General overview of an $\CAE$ used for document image binarization (input image taken from~\cite{PratikakisZBG16}). The output layer consists of the activation level assigned to each input feature (white denotes maximum activation).}
	\label{fig:hierarchy_sae}
\end{figure}

Given that we are dealing with images, the hierarchy of layers of our $\CAE$ consists of a series of convolutional plus down-sampling layers, until reaching an intermediate layer in which a meaningful representation of the input is attained (function $f$). As the down-sampling layers are applied, convolutions are able to relate parts of the image that were initially far apart. It then follows a series of convolutional plus up-sampling layers that reconstruct the image up to its initial size (function $g$). The last layer consists of a set of neurons that predict a value in the range of $[0,1]$, depending on the selectional level of the corresponding input pixel. A graphical illustration of this configuration is shown in Fig.~\ref{fig:hierarchy_sae}

\begin{figure}[tbp]
        \centering
        \begin{subfigure}[b]{0.48\textwidth}
         \includegraphics[width=\textwidth]{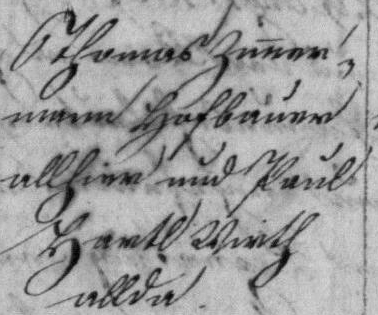}
         \caption{Input image}
        \end{subfigure} 
        \begin{subfigure}[b]{0.48\textwidth}
          \includegraphics[width=\textwidth]{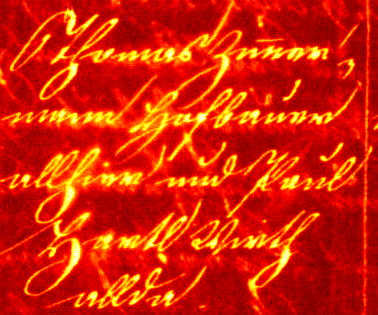}
          \caption{Activations}
        \end{subfigure}     \\ 
        \begin{subfigure}[b]{0.48\textwidth}
          \includegraphics[width=\textwidth]{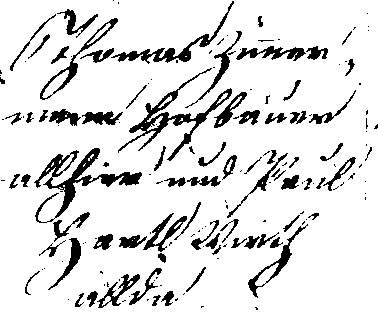}
          \caption{Thresholding}
        \end{subfigure}
         \begin{subfigure}[b]{0.48\textwidth}
          \includegraphics[width=\textwidth]{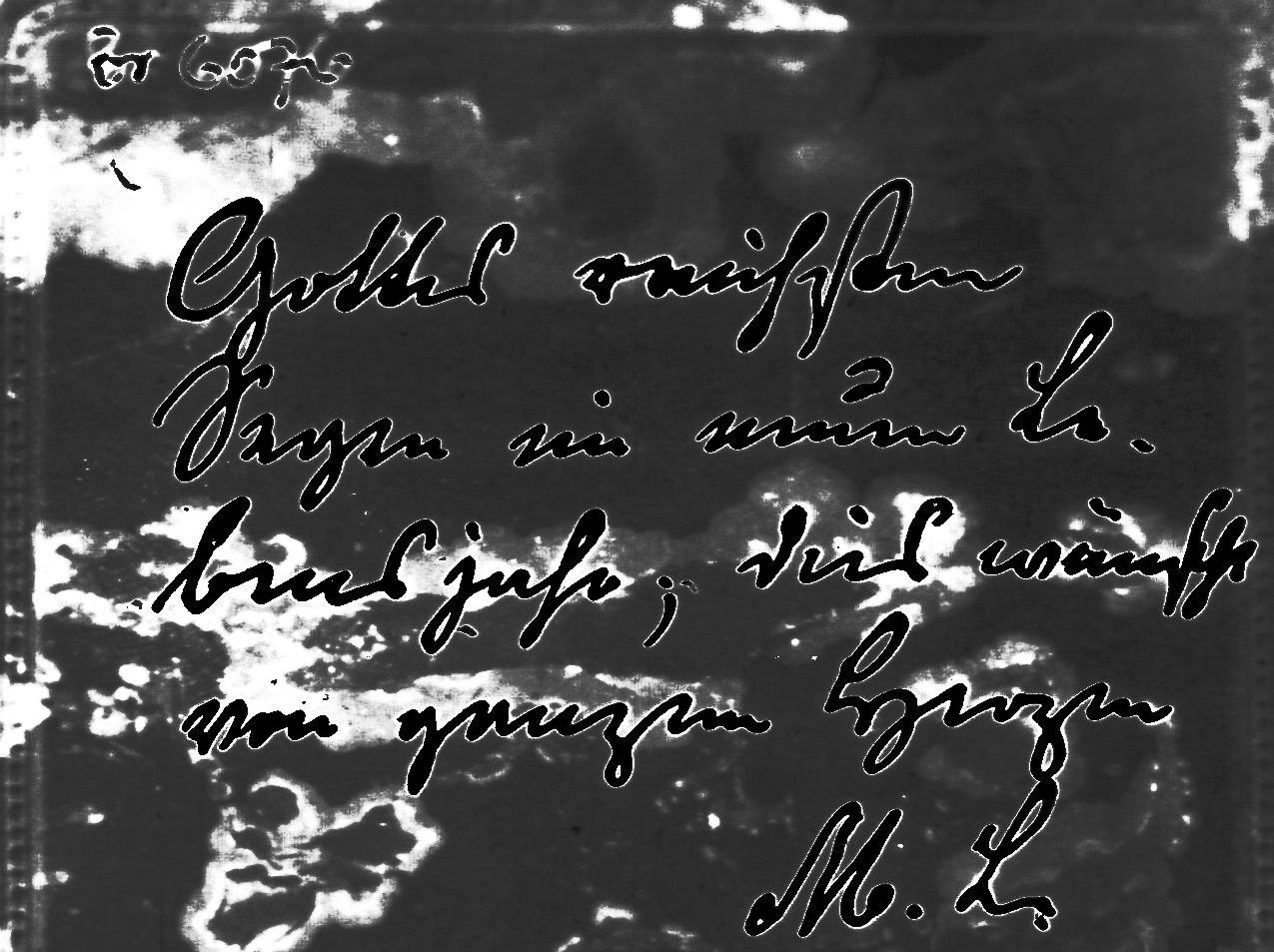}
          \caption{Diff. between activation and input}
        \end{subfigure}
        \caption{Example of binarization task 
        using an $\CAE$. A neural activation (selectional value) is obtained for each pixel of the input image, illustrated here with a heat map. A thresholding is then applied to binarize the image. The difference between the neural activations and the grayscale values of the original image is also depicted to assess the influence of the model in the process (white represents a greater difference).}
        \label{fig:binarization_cae_example}
\end{figure}

Thus, once the $\CAE$ has been properly trained, an image can be parsed through the network, after which a selection level is assigned to each input pixel. In practice, the network hardly outputs either $0$ or $1$ but an intermediate value. Therefore, a thresholding process is still necessary to convert the obtained neural activations into actual binary values. Those pixels whose selectional value exceeds a certain threshold are considered to belong to the foreground of the document image, whereas the others are labeled as background. The process is illustrated in Fig.~\ref{fig:binarization_cae_example}.  
Since the network already takes into account the context of each pixel in its internal operation, a single global threshold is sufficient. This value has to be set empirically. In our experiments, we study the robustness of the approach against the considered value.

Quite often, the size of the input document is higher than the input layer of the $\CAE$. In those cases, it is only necessary to split the input document into pieces of the size expected by the network and parse them independently.
To reconstruct the original image, it is only necessary to assemble the independent pieces, without further processing. As will be shown in the experimental analysis, the approach does not produce recurring errors or discontinuities in the regions that are close to the edges of these pieces.

\subsection{Implementation details} \label{subsec:implementation}
We have introduced previously the proposed framework to binarize documents using $\CAE$. Nevertheless, a series of implementation details need to be established to carry out this idea. 

For instance, the specific topology of an auto-encoder can be very varied. In this paper, we study three possibilities:

\begin{itemize}
\item Convolutional Auto-Encoder (CAE): traditional encoder-decoder architecture considering convolutional layers \cite{masci2011stacked}. The encoder function is implemented by  layers of convolution plus max-pooling operations, whereas the decoder function consist of layers of convolution plus up-sampling operators.
\item Stacked What-Where Auto-Encoder (SWWAE) \cite{zhao2015stacked}: in this topology the pooling layers of the encoding function produce two types of connections. The \emph{what} connection is fed to the following layer, while the \emph{where} connection is fed to the analogous layer of the decoder function. We consider the instance of the SWWAE that uses a convolutional net as encoder and a deconvolutional net \cite{ZeilerKTF10} as decoder.
\item Very deep Residual Encoder-Decoder Network (RED-Net) \cite{MaoSY16}:  this topology includes residual connections from each encoding layer to its analogous decoding layer, which facilitates convergence and leads to better results. In addition, down-sampling is performed by convolutions with stride, instead of resorting to pooling layers. Up-sampling is achieved through deconvolutional layers.
\end{itemize}

For all models, 5 layers of encoding and 5 layers of decoding are considered, and the sampling operators are fixed to $2 \times 2$. The other hyper-parameters, including the size of the input/output, will be studied empirically.

It should be remembered that, since we are implementing a selectional approach, the network outputs are one-channel images of the same size as the input whose pixel values are restricted to the range $[0,1]$. This is achieved by considering sigmoid activations in the last layer. In addition to the topology of the network, there exist hyper-parameters (such as the size of the input layer, the number of filters per layer, the size of the filter kernels, and so on) which will be studied experimentally.

Networks weights are initialized according to the so-called Xavier uniform initializer~\cite{glorot2010understanding}. The optimization of these weights is carried out by means of stochastic gradient descent~\cite{bottou2010large}, with a mini-batch size of $10$, and considering the adaptive learning rate strategy proposed by Kingma and Ba~\cite{kingma2014adam}, with its default parameterization (initial learning rate set to $0.001$). Since document binarization typically represents an imbalanced problem (more background than foreground), we make use of the loss function proposed by Pastor-Pellicer et al.~\cite{pastor2013f}, which focuses on maximizing the F-measure.  
The training process is carried out for a maximum of 200 epochs. However, we follow an \emph{early stopping} strategy, and so  the training process is stopped if the training loss does not decrease after 20 epochs. Once the process ends, the configuration of the epoch with the lowest loss is finally selected.
In addition, a slight data augmentation procedure is considered so that new synthetic samples are generated by doing random scaling and flips over the original images. The impact of this process will be evaluated empirically.

\section{Experiments} \label{sec:experiments} 
This section details the experimentation carried out to evaluate the discussed  approach. The performance of a binarization algorithm can be evaluated in several ways. For instance, if the algorithm is part of a workflow to perform a particular task, an interesting way to measure the performance is in relation to the final performance. However, this implies that the evaluation of the algorithm may not be totally fair, since it would be strongly related to the performance of the rest of the stages of that workflow.

In our case, we resort to the F-measure (Fm) to perform a direct evaluation of the binarization process. This metric is suitable for two-class problems in which the distribution of the samples is not balanced. Let TP, FP and FN stand for true positives (foreground pixels classified as foreground), false positives (background pixels classified as foreground), and false negatives (foreground pixels classified as background), respectively. Thus, the Fm can be computed as
$$
\text{Fm} = \frac{2 \, \text{TP}}{2 \, \text{TP} + \text{FP} + \text{FN}}.
$$

In the first series of experiments, we study the influence of both the chosen network topology and its hyper-parameters in order to select a suitable configuration for the document image binarization task. After that, we compare the goodness of the $\CAE$ approach against existing binarization methods. We also include further analysis of the performance, focusing on the errors made and the ability to adapt to different document types. 

The source code to reproduce our experiments is freely available. \footnote{\texttt{\url{https://github.com/ajgallego/document-image-binarization}}, under the conditions of the GNU General Public License version 3.}

\subsection{Corpora} \label{subsec:corpora}
In order to train the considered models and evaluate the performance of the algorithms, it is necessary to resort to ground-truth data. In this paper we are interested in validating the goodness of the binarization strategies in a wide range of document types. For this, we make use of corpora from different domains, namely those listed below:

\begin{itemize}
\item Latin text: the Document Image Binarization Contest (DIBCO) has been held for several years, starting from 2009~\cite{GatosNP09}. In each edition, the contest provides a series of images depicting Latin text that must be binarized, and the algorithms are evaluated based on a ground-truth created in a supervised way. In this work, we consider the 2014 (D14) \cite{ntirogiannis2014icfhr2014} and 2016 (D16) \cite{PratikakisZBG16} editions. Within these datasets, we find images with both printed and handwritten text.
\item Palm leaf: a competition on the binarization of Balinese palm leaf manuscripts has been recently organized within the 15th International Conference on Frontiers in Handwriting Recognition \cite{burie2016icfhr2016}. The contest provided two different corpora depending on how the ground-truth was built (PL-I and PL-II).
\item Persian documents: the set of documents from the Persian Heritage Image Binarization Competition (PHI) \cite{ayatollahi2013persian} are considered here as a way of evaluating the algorithms in a different text domain.
\item Music scores: the last set of corpora consist of document images depicting music scores, which includes both music notation and text (lyrics). We specifically make use of high-resolution scans of two different old music documents: Salzinnes Antiphonal Manuscript (SAM) and Einsiedeln Stiftsbibliothek (ES), whose binarization has been studied previously~\cite{calvo17binarization}.
\end{itemize}

\begin{figure}[tbp]
        \centering
        \begin{subfigure}[b]{0.32\textwidth}
         \includegraphics[width=\textwidth]{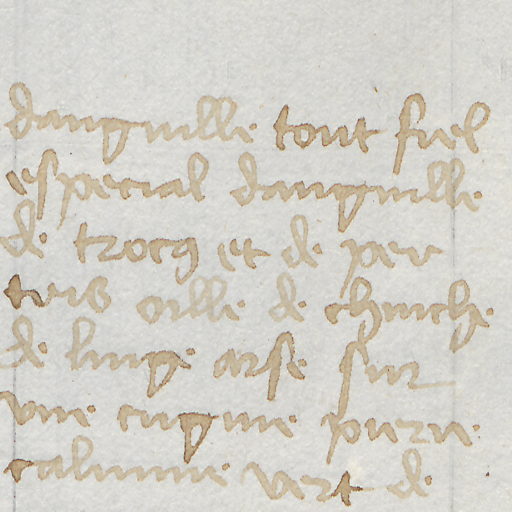}
         \caption{D14}
        \end{subfigure} 
        \begin{subfigure}[b]{0.32\textwidth}
          \includegraphics[width=\textwidth]{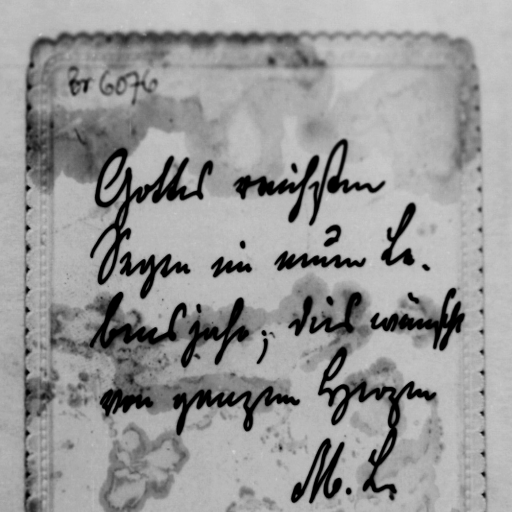}
          \caption{D16}
        \end{subfigure}     
        \begin{subfigure}[b]{0.32\textwidth}
          \includegraphics[width=\textwidth]{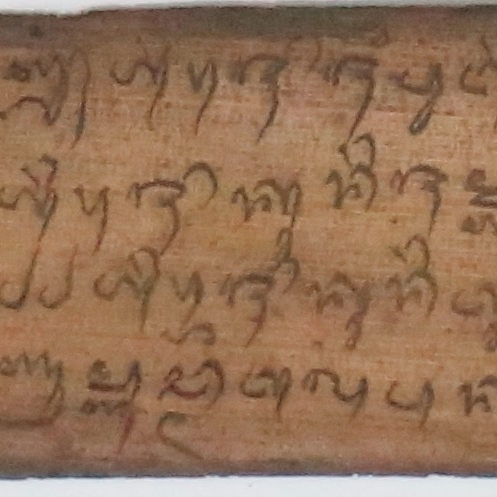}
          \caption{PL-I \& PL-II}
        \end{subfigure}
        \\
        \begin{subfigure}[b]{0.32\textwidth}
         \includegraphics[width=\textwidth]{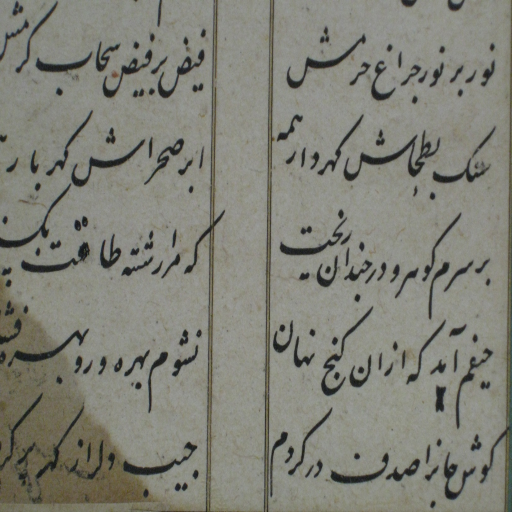}
         \caption{PHI}
        \end{subfigure} 
        \begin{subfigure}[b]{0.32\textwidth}
          \includegraphics[width=\textwidth]{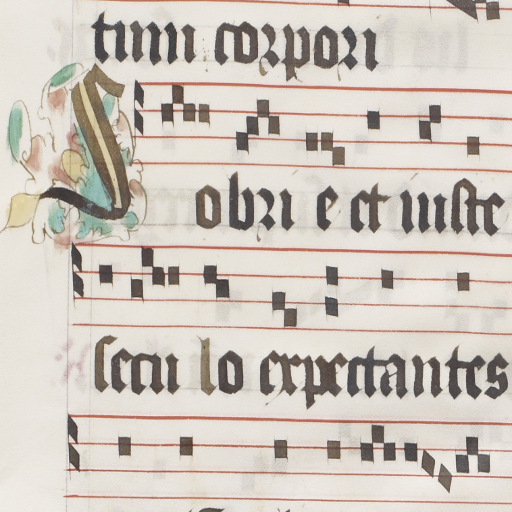}
          \caption{SAM}
        \end{subfigure}     
        \begin{subfigure}[b]{0.32\textwidth}
          \includegraphics[width=\textwidth]{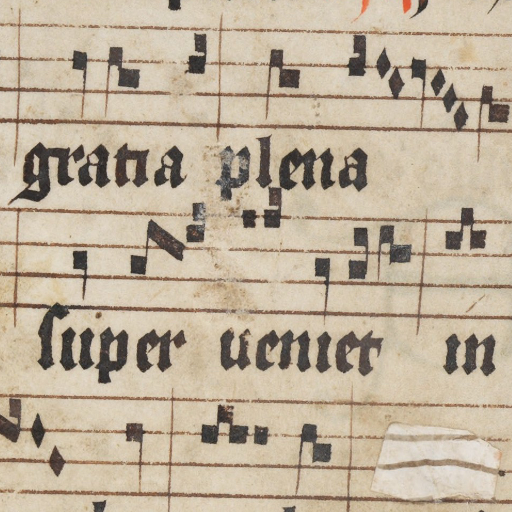}
          \caption{ES}
        \end{subfigure}
\caption{Examples of document patches from the considered corpora.}
\label{tab:data_examples}
\end{figure}

As example, a patch of document from each dataset is depicted in Fig~\ref{tab:data_examples}.

Given that the $\CAE$ needs training data, we split the corpora in train and test partitions. The former are used for training and validating the models, while the latter are used for computing the evaluation metrics. In the case of DIBCO data (D14 and D16), we use in each case all the images from the rest of the contest editions as training set. PL-I and PL-II are already provided with their train and test partitions. In the case of PHI, SAM, and ES, we randomly split the corpora into train and test, with 80\% and 20\% of the available documents, respectively. A summary of the statistics of each corpus is given in Table \ref{tab:corpora}, including the number of documents and the approximate number of pixels (in millions) of each partition.

\begin{table}[ht]
\centering
\renewcommand{\arraystretch}{1.4}
\begin{tabular}{llcclcc}

\toprule[1pt]
\multirow{2}{*}{Dataset} & \multirow{2}{*}{Type} & \multicolumn{2}{c}{Documents} & & \multicolumn{2}{c}{Million pixels} \\ \cline{3-4} \cline{6-7}
 & & Train           & Test & & Train     & Test \\ \hline
D14         & Latin text            & 76 & 10 & & 85 & 10  \\
D16         & Latin text            & 76 & 10 & & 82 & 13 \\ 
PL-I        & Palm leaf             & 50 & 50 & & 125 & 125 \\ 
PL-II       & Palm leaf             & 50 & 50 & & 125 & 125 \\ 
PHI         & Persian               & 12 & 3 & & 16 & 2 \\ 
ES          & Music score           & 8 & 2 & & 162 & 41 \\ 
SAM         & Music score           & 8 & 2 & & 131 & 33 \\
\bottomrule[1pt]
\end{tabular}
\caption{Corpora considered in our experiments.}
\label{tab:corpora}
\end{table}

\subsection{Hyper-parameter selection} \label{subsec:model_selection}
As discussed earlier, this work focuses on the binarization approach of documents using $\CAE$. The main emphasis is on the approach itself and not so much on finding the best configuration of the model. However, it is obvious that we need to establish a specific configuration that allows us to evaluate the capabilities of the approach. This is why we want to study how the different topologies and hyper-parameters of the models affect the performance. 

To carry out this experiment, we are going to restrict ourselves to use only D16 dataset. The reason is twofold: on the one hand, Latin text binarization from DIBCO competitions is a standard benchmark for evaluating binarization strategies, so D16 is a good representative of the task at hand; on the other hand, we do not want our final results depending on an excessive tuning of the model that best fits. Our premise is that the goodness of the $\CAE$ approach lies in its formulation, and not in selecting the optimum configuration for every case. We therefore assume that the best configuration in this study will be a good representative of the $\CAE$ approach for document image binarization. 

Since the general configuration of the models has already been specified previously (see Section \ref{subsec:implementation}), we only study three hyper-parameters: the size of the input/output window (patch of the document that is processed in a single step), the number of convolutional filters in each layer, and the kernel size of the convolutional filters.

\begin{figure}[ht]
\centering
	\includegraphics[width=.7\columnwidth]{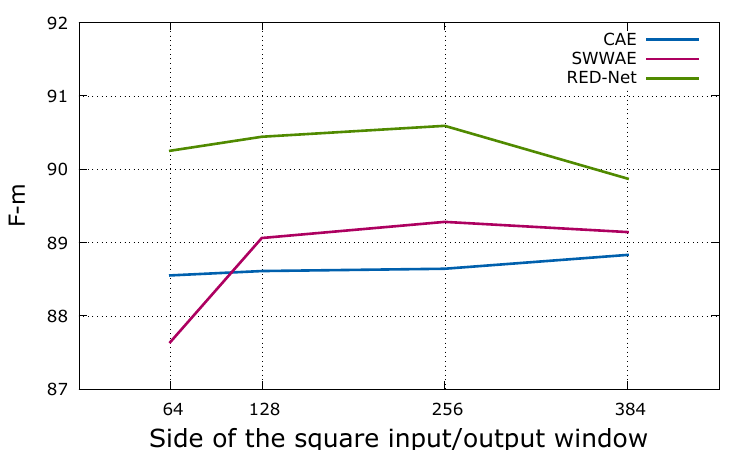}
	\caption{Influence in terms of Fm of the size of the square window with respect to each considered $\CAE$ model.}
	\label{fig:wsize}
\end{figure}

The first aspect to consider is the size of the window, fixing $64$ filters per convolution and kernel size of $3 \times 3$. In order to reduce the search space, we have restricted ourselves to square windows of sizes $64\times64$, $128\times128$, $256\times256$, and $384\times384$. Figure~\ref{fig:wsize} shows that this parameter can strongly influence the performance of the model. For instance, it is observed that RED-Net achieves the highest Fm when considering windows of size $256\times256$, degrading significantly its performance with windows of size $384\times384$. Also, the SWWAE obtains the worst performance with $64 \times 64$ windows, but its results increase noticeably with bigger windows. The CAE model reports a more robust behavior with respect to this parameter, although its results are generally below the other configurations.

\begin{figure}[ht]
\centering
	\includegraphics[width=.7\columnwidth]{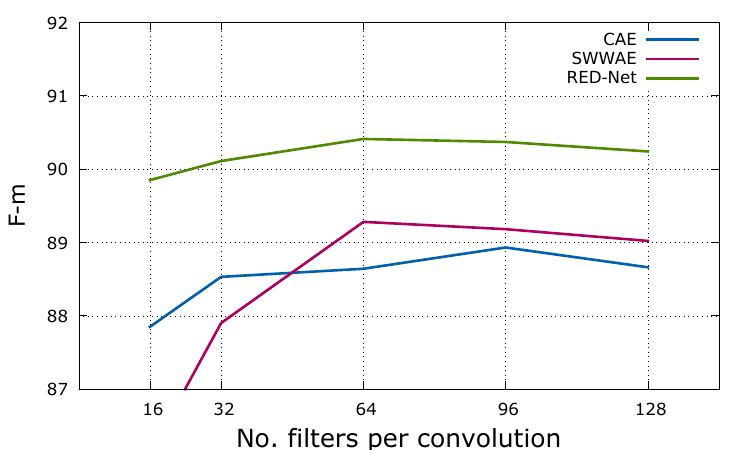}
	\caption{Influence in terms of Fm of the number of filters per convolution with respect to each considered $\CAE$ model.}
	\label{fig:fsize}
\end{figure}

The second hyper-parameter to study is the number of filters per convolutional layer. In spite of knowing that it is not necessarily optimal, let us force that all convolutional layers consist of the same number of filters so that this study does not become huge. Specifically, we consider $16$, $32$, $64$, $96$, and $128$ filters per layer. In this case, we keep the kernel size to $3\times3$ but set the input size to the best value obtained in the previous experiment for each model: $384\times384$ for CAE, and $256\times256$ for SWWAE and RED-Net. Figure~\ref{fig:fsize} shows the curves of this experiment. It is observed that the number of filters has a smaller influence on RED-Net but greater on SWWAE and CAE. In the SWWAE model, it is relevant because it varies from less than $87$ (16 filters) to $89.3$ (64 filters) of Fm. Again, the best value is attained by RED-Net, specifically with $64$ filters.

\begin{figure}[ht]
\centering
	\includegraphics[width=.7\columnwidth]{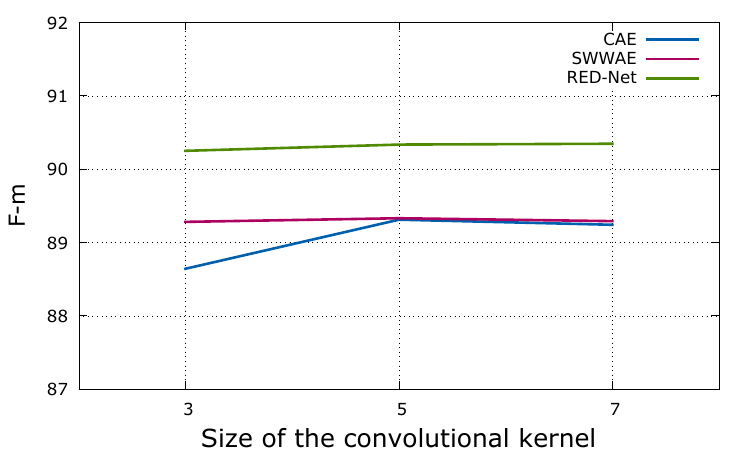}
	\caption{Influence in terms of Fm of the size of the convolutional kernel with respect to each considered $\CAE$ model.}
	\label{fig:ksize}
\end{figure}

The last model hyper-parameter to study is the kernel size of the convolutional operators. As in the previous case, we consider the same number for all layers: square kernels of size $3\times3$, $5\times5$, or $7\times7$. The rest of the studied parameters are set according to the best configuration of each model in the previous experiments. As illustrated in Fig.~\ref{fig:ksize}, both RED-Net and SWWAE are hardly sensitive to this parameter. It can be seen that their curves are almost straight, albeit with a negligible improvement as the value increases in the case of RED-Net. The kernel size seems to be somewhat relevant for CAE, as its performance slightly improves with values higher than $3\times3$. 

Generally, it can be seen that the $\CAE$ approach itself is quite robust beyond configurations and hyper-parameters, as the results only range from $88$ to $91$ of Fm. In particular, the RED-Net model with $256 \times 256$, $64$ filters, and $7 \times 7$ kernels attains the best performance amongst the evaluated possibilities. However, we finally selected a kernel size of $5 \times 5$ given that $7 \times 7$ kernels implies a much larger network with an insignificant improvement.

A summary of the evaluation carried out can be found in Table \ref{tab:final_configuration_preliminary}, where we present the final configuration of the neural models as well as their obtained result, in this preliminary experiment. Although the differences are not very remarkable, we can observe that RED-Net slightly outperforms the other neural models.
From here on, we shall consider this full configuration as a representative of the $\CAE$ approach for document binarization.

\begin{table}[ht]
\centering
\begin{tabular}{lccc}
\hline
\textbf{Model:}                 & \textbf{CAE}   & \textbf{SWWAE} & \textbf{RED-Net} \\ \hline
\textbf{Input image size (px):} & 384$\times$384 & 256$\times$256 & 256$\times$256   \\
\textbf{Number of layers:}      & 5+5            & 5+5            & 5+5              \\
\textbf{Filters per layer:}     & 96             & 64             & 64               \\
\textbf{Kernel size:}           & 5$\times$5     & 5$\times$5     & 5$\times$5       \\ \hline
\textbf{Fm:} 					&  89.3	& 89.3	 & 90.3           \\ \hline
\end{tabular}
\caption{Configuration and final result in the preliminary evaluation of the neural models.}
\label{tab:final_configuration_preliminary}
\end{table}

\begin{figure}[ht]
\centering
	\includegraphics[width=.7\columnwidth]{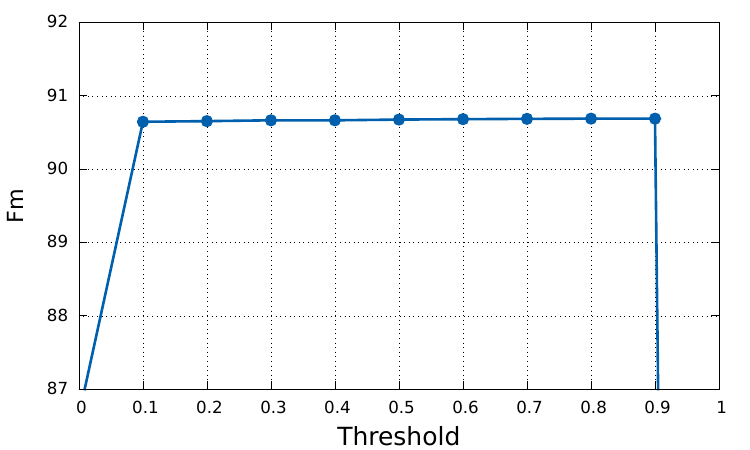}
	\caption{Influence in terms of Fm of the threshold in the selected $\CAE$ model.}
	\label{fig:threshold_evaluation}
\end{figure}

After selecting this model, we also studied the influence of the threshold used to convert the selectional values (neural activations) into the expected binary output. We measured the Fm obtained for different threshold values, within the range of $[0.1,0.9]$, as shown in Fig.~\ref{fig:threshold_evaluation}.

It is observed that the variability of the performance with respect to this parameter is negligible, hardly varying the results except for the extreme values (0 or 1). This fact reinforces the robustness of the activations predicted by the $\CAE$ because they tend to be close to either $0$ or $1$, thereby decreasing the importance of the selected threshold. Given the limited influence, we will simply use a value of $0.5$ as a threshold in the subsequent experiments.

\begin{figure}[ht]
\centering
	\includegraphics[width=.7\columnwidth]{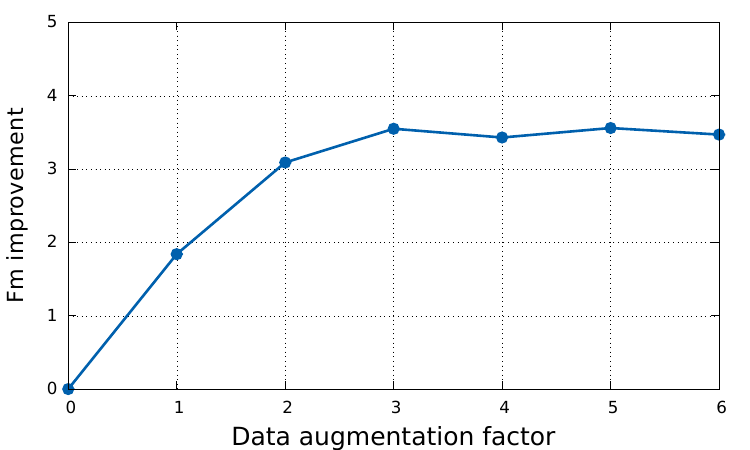}
	\caption{Improvement in terms of Fm of the data augmentation factor in the selected $\CAE$ model.}
	\label{fig:daug}
\end{figure}

Finally, we introduced in the previous section the consideration of data augmentation by means of random flips and vertical scaling on the original images, using random factors within the range $[0.5, 1.5]$. The number of times that this process is performed has been evaluated empirically, oscillating between 0 (no augmentation) and 6 new images from each of those already available. In each generated sample, both effects are used simultaneously. The impact of this process is depicted in Fig. \ref{fig:daug}, in which the improvement of this process in terms of Fm is measured with respect to the number of samples generated from each original image (data augmentation factor).

It can be seen that data augmentation is indeed effective, as it leads to improvements of around 4\% of Fm. However, the effect produced by the augmentation factor quickly stabilizes in 3, after which the improvement is variable. Thus, taking into account that a greater number of samples is detrimental to the computational cost of the training process, we shall consider a factor of 3 for data augmentation in the remaining experiments.

\subsection{Comparative assessment}
In this section we compare the performance of the $\CAE$ approach against other binarization algorithms, including a baseline reference as well as classical and state-of-the-art algorithms. Specifically, we consider the following methods:

\begin{itemize}
\item Baseline: Otsu~\cite{otsu1975threshold} and Niblack~\cite{niblack1985introduction}. Although these methods are old, they still remain as good baseline references.
\item Classic: Sauvola~\cite{sauvola2000adaptive}, Wolf et al.~\cite{wolf02text}, and Gatos et al.~\cite{gatos06adaptive}. These strategies have been traditionally used to binarize document images in many existing workflows.
\item State of the art: Su et al.~\cite{su2013robust}, Sauvola MS (Lazzara and Geraud \cite{lazzara2014efficient}), Howe~\cite{howe2013document}, and Kliger and Tal~\cite{PratikakisZBG16}. Recent methods that constitute the state of the art for document image binarization.
\end{itemize}

In addition, we include a pixel-wise CNN approach, as a representative of learning-based binarization with neural networks. All these strategies have been introduced previously in more detail (cf. Section~\ref{sec:background}). 
The implementations of these algorithms come from those provided by the authors of the corresponding publications. The exceptions are Otsu, Niblack and Sauvola, for which we used the \emph{scikit-image} Python package implementation, \footnote{http://scikit-image.org/} and the pixel-wise CNN approach, which we manually implemented following the specifications detailed in~\cite{calvo17binarization}.

Furthermore, there exist two possibilities to train the selected $\CAE$ model that we analyze empirically: (i) training the model only with the specific train partition of each dataset ($\CAE$ specific); and (ii) training the model with the train partitions of all datasets ($\CAE$ global). Thus, we evaluate if the $\CAE$ benefits more from the suitability of the data or from its quantity.

Table~\ref{tab:results_comparison} presents the figures obtained in terms of Fm for each corpus. An initial remark to begin with is that there is a noticeable difference amongst the performance attained by the different strategies in each dataset, reinforcing our idea of including as many types of document as possible in these experiments.

\begin{table}
\centering
\renewcommand{\arraystretch}{1.4}
\resizebox{1\textwidth}{!}{
\begin{tabular}{lcccccccc}
\toprule[1pt]
			 & \multicolumn{7}{c}{Dataset} \\ \cline{2-8}
Method 	 	  	& D14 & D16 & PL-I & PL-II & PHI & ES & SAM & Avg. \\ \hline
Otsu~\cite{otsu1975threshold} 			& 91.56 & 73.79 & 34.92 & 34.75 & 90.77 & 74.32 & 79.72 & 68.55 \\
Niblack~\cite{niblack1985introduction} 		& 22.26 & 16.7  & 16.25 & 16.15 & 25.70 & 34.3  & 31.96 & 23.33 \\
Sauvola~\cite{sauvola2000adaptive} 		& 77.08 & 82.00	& 9.94  & 9.75  & 81.64 & 61.69 & 65.14 & 55.32 \\
Wolf et al.~\cite{wolf02text}		& 90.47 & 81.76 & 30.72 & 30.30 & \underline{91.84} & 67.02 & 77.00 & 67.02 \\
Gatos et al.~\cite{gatos06adaptive}    & 91.97 & 74.97 & 19.75 & 19.88 & 91.69 & 74.63 & 80.11 & 64.71 \\
Sauvola MS~\cite{lazzara2014efficient} 		& 87.86 & 65.04 & 42.76 & 42.41 & 91.39 & 74.00 & 79.79 & 69.03 \\
Su et al.~\cite{su2013robust}		& \underline{95.14} & 90.27 & 48.44 & 48.04 & 89.27 & 75.61 & 81.55 & \underline{75.48} \\
Howe~\cite{howe2013document} 			& 90.00 & 80.64 & 46.49 & 45.77 & 88.97 & 76.03 & 80.52 & 72.63 \\
Kliger and Tal~\cite{PratikakisZBG16} 	& 95.00 & \underline{90.48} & 46.58 & 45.88 & 89.67 & \underline{76.88} & \underline{82.09} & 75.22 \\
CNN~\cite{calvo17binarization} 			& 81.23 & 54.58 & \underline{51.47} & \underline{51.09} & 88.53 & 75.81 & 81.05 & 69.11  \\ \hline
$\CAE$ specific		& \textbf{98.05} & \textbf{91.65} & \textbf{69.65} & \textbf{69.13} & \textbf{93.82} & \textbf{78.45} & \textbf{83.13} & \textbf{83.41} \\
$\CAE$ global	& 89.12 & 85.27 & \textbf{63.74} & \textbf{63.21} & 71.63 & \textbf{77.11} & 81.38 & \textbf{75.92} \\
  \bottomrule[1pt]
\end{tabular}
}
\caption{Performance comparison in terms of Fm of the $\CAE$ approach against existing binarization strategies. Values underlined represent the best performance in each dataset by existing algorithms (state of the art); values in bold represent the results of the $\CAE$ approach that outperform the state of the art.}
\label{tab:results_comparison}
\end{table}

In general, baseline and classical methods get worse performance, with some exceptional cases that depend on the method and the dataset. This can be seen in the irregularity of methods such as Gatos et al. or Otsu, which achieve good results in D14, PHI, ES or SAM but behave poorly in P-I and P-II. Furthermore, it is observed that state-of-the-art methods obtain a higher average performance (\emph{Avg.} column) than the previous ones. In particular, the method proposed by Kliger and Tal obtains the best result among the existing methods in three of the seven datasets considered, although the method proposed by Su et al. attains the best results on average. The learning-based approach with CNN depicts a quite irregular behavior, which ranges from being the best method within the existing ones in two datasets (P-I and P-II) to not outperforming the specialized level of other methods with text documents. However, it is closer to the classical ones than to the considered state-of-the-art methods, on average.

Regarding the $\CAE$ approach, it is observed that it is better to train the model with specialized data ($\CAE$ specific) since results are always superior to the case of training with all available data ($\CAE$ global). In the former case, the performance is able to improve all the figures obtained by existing algorithms, whereas in the latter many of them are improved as well. It can be seen that the improvement is especially remarkable in those datasets where the existing methods find more difficulties (\emph{ie.}, P-I and P-II). Focusing on the average case, the studied approach is able to increase the state of the art from $75.48$ to $83.41$.

Although it is better to train the with specific data, it is important to emphasize that the $\CAE$ global also obtains highly competitive results. In fact, it improves the performance of existing algorithms in many datasets, and also in the average case (from $75.48$ to $75.92$ of Fm). This represents an excellent result considering that the model is learning with data of very different typology, and it is able to generalize correctly. This fact is also important in practice because it may reduce the need for manuscript-specific ground-truth data in a new document type, because the global model seems to behave well in general.

Note that errors may be of different relevance depending on where they occur. That is why we will also  evaluate the performance as regards a pseudo Fm (p-Fm) specially designed for binarization problems. Such p-Fm was introduced in~\cite{ntirogiannis2013performance}, and it is similar to the classical Fm but weighing the relevance of each pixel depending whether it is close to stroke boundaries. We show in Table \ref{tab:fps_comparison} the previous comparative but evaluating the performance with this metric. 
It can be verified that the analysis performed above still holds: although some algorithms have modified their order with respect to the best result with existing algorithms, the $\CAE$ achieves the best p-Fm performance in all datasets when it is trained with specific data. In addition, the global $\CAE$ is highly competitive, as its p-Fm is higher than those of previous approaches, on average. However, it can be observed that the improvement introduced by the $\CAE$ is less pronounced as regards p-Fm. It is possible that this difference is caused by training the $\CAE$ to optimize the classical Fm, instead of p-Fm.

\begin{table}
\centering
\renewcommand{\arraystretch}{1.4}
\resizebox{1\textwidth}{!}{
\begin{tabular}{lcccccccc}
\toprule[1pt]
                & \multicolumn{7}{c}{Dataset}                                                                                          & \multicolumn{1}{l}{} \\ \cline{2-8}
Method          & \textbf{D14}   & \textbf{D16}   & \textbf{PL-I}  & \textbf{PL-II} & \textbf{PHI}   & \textbf{ES}    & \textbf{SAM}   & Avg.                 \\ \hline
Otsu~\cite{otsu1975threshold}            & 95.69          & 71.85          & 31.82          & 31.63          & 91.86          & 67.33          & 74.08          & 66.32                \\
Niblack~\cite{niblack1985introduction}         & 20.96          & 15.93          & 14.85          & 14.76          & 23.62          & 31.67          & 30.26          & 21.72                \\
Sauvola~\cite{sauvola2000adaptive}         & 85.31          & 82.45          & 11.37          & 11.06          & 90.98          & 55.72          & 61.65          & 56.94                \\
Wolf et al.~\cite{wolf02text}     & 93.91          & 80.59          & 32.29          & 31.69          & 93.68          & 60.82          & 72.11          & 66.44                \\
Gatos et al.~\cite{gatos06adaptive}    & 95.39          & 73.25          & 18.48          & 18.58          & \underline{94.62}    & 67.68          & 74.53          & 63.22                \\
Sauvola MS~\cite{lazzara2014efficient}      & 92.08          & 62.93          & 44.04          & 43.43          & 94.03          & 66.97          & 74.18          & 68.24                \\
Su et al.~\cite{su2013robust}       & 96.67          & 90.76          & 46.50          & 46.03          & 87.97          & 68.71          & 75.84          & 73.21                \\
Howe~\cite{howe2013document}            & 91.82          & 80.54          & 45.51          & 44.64          & 89.28          & 69.08          & 74.88          & 70.82                \\
Kliger and Tal~\cite{PratikakisZBG16}  & \underline{97.92}    & \underline{91.00}    & 45.27          & 44.43          & 90.05          & \underline{69.88}    & \underline{76.36}    & \underline{73.56}          \\
CNN~\cite{calvo17binarization}             & 78.85          & 53.56          & \underline{49.52}    & \underline{49.07}    & 88.70          & 69.06          & 75.65          & 66.34                \\ \hline
$\CAE$ specific & \textbf{98.47} & \textbf{93.71} & \textbf{68.54} & \textbf{67.94} & \textbf{94.66} & \textbf{70.48} & \textbf{76.65} & \textbf{81.49}       \\
$\CAE$ global   & 87.85          & 86.95          & \textbf{63.45} & \textbf{62.72} & 73.61          & 68.94          & 74.68          & \textbf{74.03}       \\ 
\bottomrule[1pt]
\end{tabular}
}
\caption{Performance comparison in terms of p-Fm of the $\CAE$ approach against existing binarization strategies. Values underlined represent the best performance in each dataset by existing algorithms (state of the art); values in bold represent the results of the $\CAE$ approach that outperform the state of the art.}
\label{tab:fps_comparison}
\end{table}

For all the above, the goodness of the $\CAE$ approach is demonstrated given that it represents an effective document image binarization method based on machine learning that is able to outperform the state of the art.

\subsection{Performance analysis}
In this section the performance of the approach is analyzed, with special attention to the errors produced.

\begin{figure}[!ht]
\centering
\renewcommand{\arraystretch}{1.4}
\resizebox{1\textwidth}{!}{
\begin{tabular}{|c|c|c|c|}
\hline
 & D16 & PL-I & ES \\ \hline
\begin{minipage}[c]{0.1\textwidth}
\centering
Source
\end{minipage} & 
\begin{minipage}[c]{0.3\textwidth} 
\includegraphics[width=1\columnwidth]{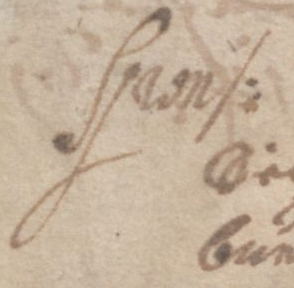} 
\end{minipage} & 
\begin{minipage}[c]{0.3\textwidth} 
\includegraphics[width=1\columnwidth]{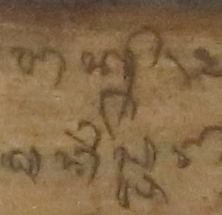} 
\end{minipage} & 
\begin{minipage}[c]{0.34\textwidth} 
\includegraphics[width=1\columnwidth]{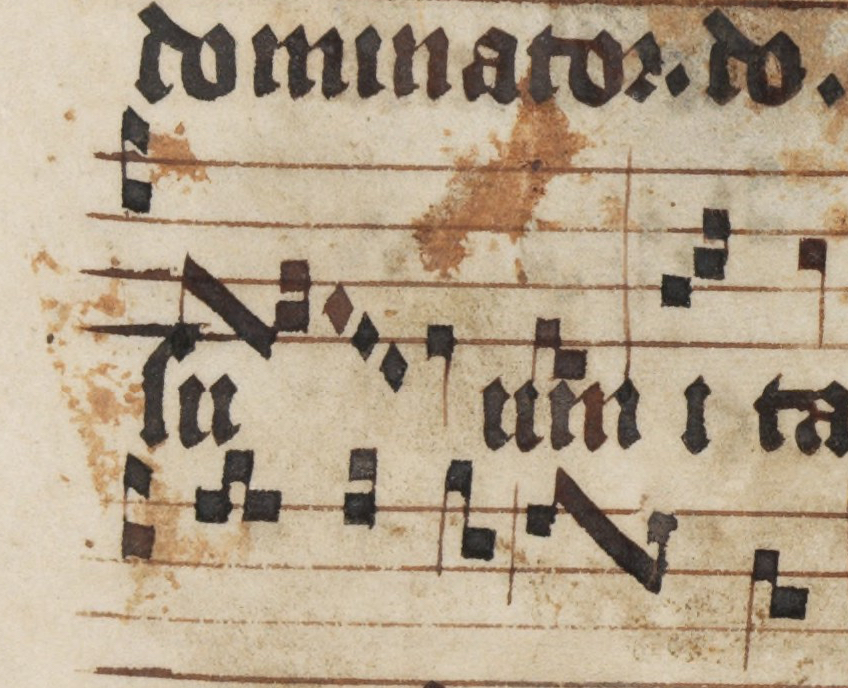} 
\end{minipage} 
\\ \hline
\begin{minipage}[c]{0.1\textwidth}
\centering
$\CAE$
\end{minipage} & 
\begin{minipage}[c]{0.3\textwidth} 
\includegraphics[width=1\columnwidth]{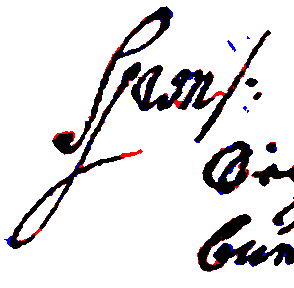} 
\end{minipage} & 
\begin{minipage}[c]{0.3\textwidth} 
\includegraphics[width=1\columnwidth]{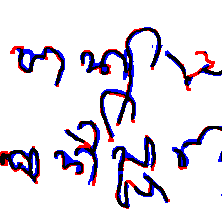} 
\end{minipage} & 
\begin{minipage}[c]{0.34\textwidth} 
\includegraphics[width=1\columnwidth]{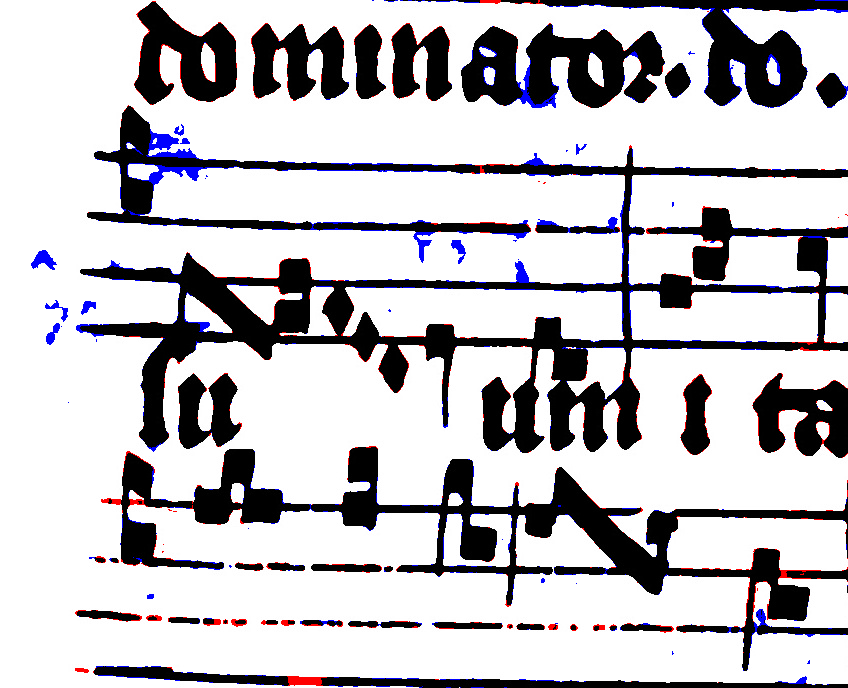} 
\end{minipage} 
\\ \hline
\begin{minipage}[c]{0.1\textwidth}
\centering
State of the art
\end{minipage} & 
\begin{minipage}[c]{0.3\textwidth} 
\includegraphics[width=1\columnwidth]{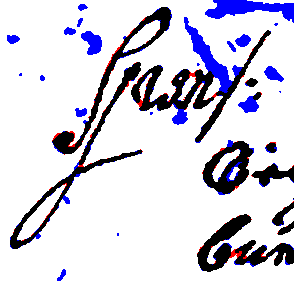} 
\end{minipage} & 
\begin{minipage}[c]{0.3\textwidth} 
\includegraphics[width=1\columnwidth]{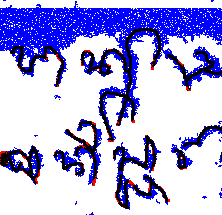} 
\end{minipage} & 
\begin{minipage}[c]{0.34\textwidth} 
\includegraphics[width=1\columnwidth]{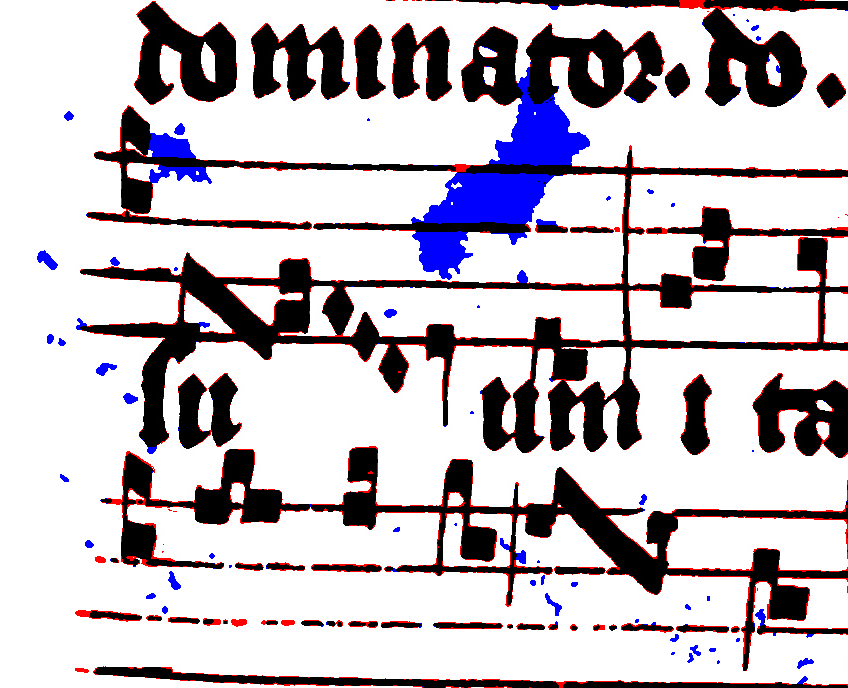} 
\end{minipage} 
\\ \hline
\begin{minipage}[c]{0.1\textwidth}
\centering
Su et al.
\end{minipage} & 
\begin{minipage}[c]{0.3\textwidth} 
\includegraphics[width=1\columnwidth]{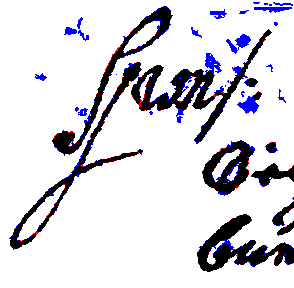} 
\end{minipage} & 
\begin{minipage}[c]{0.3\textwidth} 
\includegraphics[width=1\columnwidth]{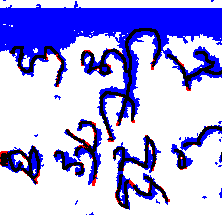} 
\end{minipage} & 
\begin{minipage}[c]{0.34\textwidth} 
\includegraphics[width=1\columnwidth]{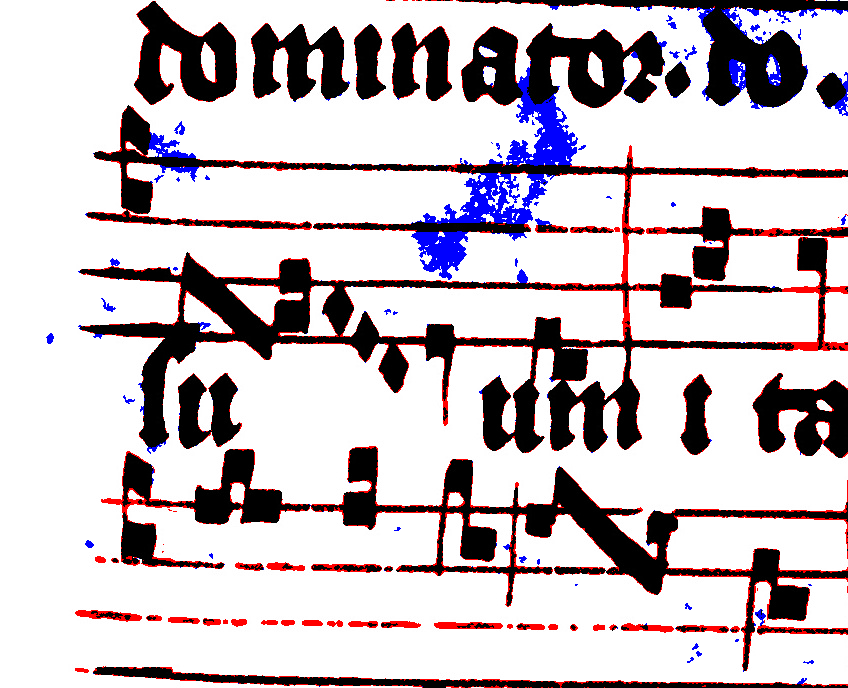} 
\end{minipage} 
\\ \hline

\end{tabular}
}
\caption{Qualitative reference of the errors produced when binarizing documents of D16, PL-I, and ES with $\CAE$, the state of the art, and Su et al. Coloring: foreground pixels identified as foreground in black, background pixels identified as background in white, foreground pixels identified as background in red, and background pixels identified as foreground in blue.}
\label{tab:error_output}
\end{figure}

First, it is interesting to show a qualitative reference of the errors produced by the $\CAE$-based binarization. For this, we have selected datasets of different types with the highest error, namely D16, PL-I and ES. Figure~\ref{tab:error_output} shows a series of representative examples of the performance obtained in these corpora, as well as where and how errors have occurred. For the sake of comparison, we include the state of the art in each dataset (Kliger and Tal in D16 and ES; CNN in PL-I), as well as that of Su et al. because of reporting the best average.

This table illustrates the location of the improvement obtained by the $\CAE$ in comparison to existing methods. Generally, it is much more effective when discarding false negatives (blue pixels), while maintaining a similar level in the detection of true positives (black pixels). That is, the approach is especially beneficial when dealing with the background of complex documents. 

Furthermore, given the underlying operation of the $\CAE$, it seems obvious to think that the errors may occur more often in the regions of the image that are closer to the edges of the considered input windows, since it is where there is less context of the neighborhood. To study this possibility, we computed a heat map of the the position of the errors with regard to the output window. We observed that errors are distributed without any defined pattern in most of the datasets, with some exceptions for palm leaf and music score documents. That is why we study the general case with D14, and the particular cases with PL-I and SAM as representatives of palm leaf and music scores, respectively.

\begin{figure}[!ht]
\centering
\renewcommand{\arraystretch}{1.7}
\resizebox{1\textwidth}{!}{
\begin{tabular}{|c|c|c|c|}
\hline
 & D14 & PL-I & SAM \\ \hline
\begin{minipage}[c]{0.2\textwidth}
\centering
$\CAE$ error heat map
\end{minipage} & 
\begin{minipage}[c]{0.3\textwidth} 
\includegraphics[width=1\columnwidth]{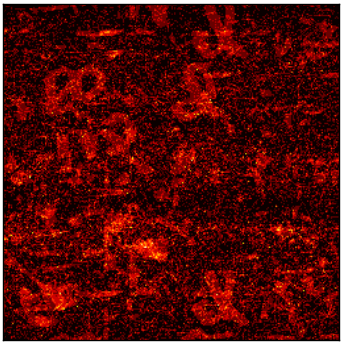} 
\end{minipage} & 
\begin{minipage}[c]{0.3\textwidth} 
\includegraphics[width=1\columnwidth]{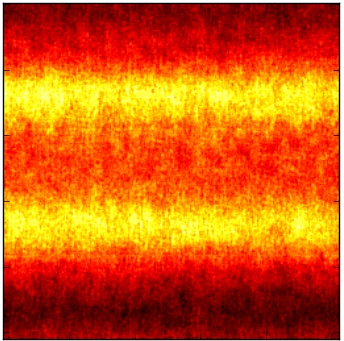} 
\end{minipage} & 
\begin{minipage}[c]{0.3\textwidth} 
\includegraphics[width=1\columnwidth]{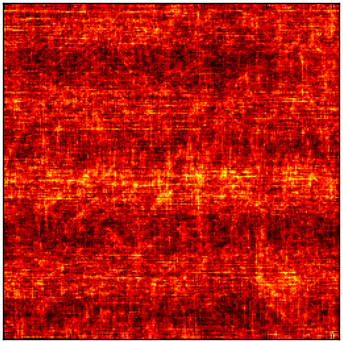} 
\end{minipage} 
\\ \hline
\begin{minipage}[c]{0.2\textwidth}
\centering
Ground-truth foreground heat map
\end{minipage} & 
\begin{minipage}[c]{0.3\textwidth} 
\includegraphics[width=1\columnwidth]{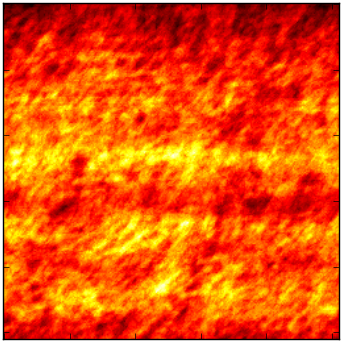} 
\end{minipage} & 
\begin{minipage}[c]{0.3\textwidth} 
\includegraphics[width=1\columnwidth]{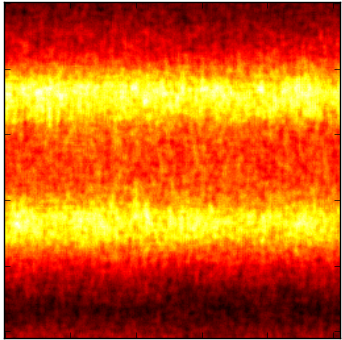} 
\end{minipage} & 
\begin{minipage}[c]{0.3\textwidth} 
\includegraphics[width=1\columnwidth]{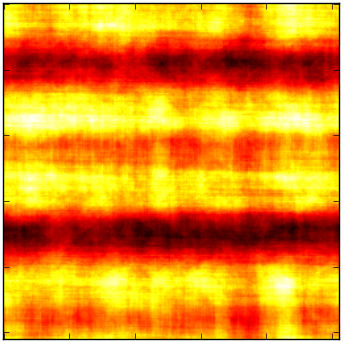} 
\end{minipage} 
\\ \hline
\begin{minipage}[c]{0.2\textwidth}
\centering
Example of ground-truth window
\end{minipage} & 
\begin{minipage}[c]{0.3\textwidth} 
\includegraphics[width=1\columnwidth]{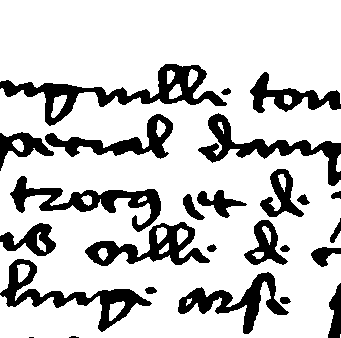} 
\end{minipage} & 
\begin{minipage}[c]{0.3\textwidth} 
\includegraphics[width=1\columnwidth]{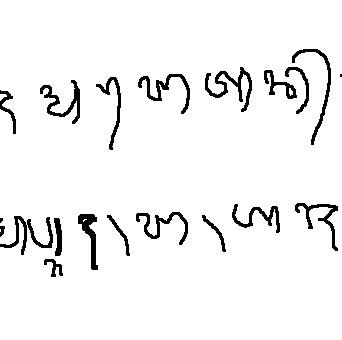} 
\end{minipage} & 
\begin{minipage}[c]{0.3\textwidth} 
\includegraphics[width=1\columnwidth]{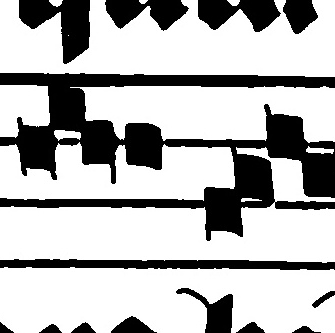} 
\end{minipage} 
\\ \hline
\end{tabular}
}
\caption{Heat map (\%) of the errors ($\CAE$ error heat map) and the foreground ground-truth pixels (ground-truth foreground heat map) with respect to positions within the  input/output window. An example of ground-truth output window is also included.}
\label{tab:heatmaps}
\end{figure}

Graphical results of this study are shown in Fig.  \ref{tab:heatmaps}. In addition to the aforementioned heat maps (error heat map), we have included another heat map that illustrates the foreground positions in the ground-truth (ground-truth heat map) and a representative ground-truth output window.

As can be observed in all cases, the areas of the image where the errors are concentrated do not have a direct relationship with their position within the window, but with the regions where the ink pixels are concentrated in the corresponding dataset. In the case of D14, as ink is distributed evenly across all positions, the error map does not depict any regular pattern. In the case of PL-I, the window is large enough to hold two text lines, so the ink map shows these two lines; accordingly, the error also seems to draw these two lines because errors are mainly produced when detecting ink (as studied previously). The case of SAM is similar: the ink map shows the four staff lines that fit in an input window, with a concentration of music symbols in the top ones. This is somewhat depicted in the error heat map as well; however, the error obtained by the $\CAE$ in this dataset is lower and, therefore, the error heat map is not so intense.

\subsection{Cross-document adaptation}
We have seen previously that an $\CAE$ is able to improve state-of-the-art binarization when the training set contains samples of the application domain. We study in this section the performance of the $\CAE$ when it is applied to document types of which no examples are provided during training.

This experiment is performed by checking how a model specifically trained for one  dataset behaves in the rest of the corpora. Table \ref{tab:domain_adaptation} shows this study, in which rows represent the training partitions and columns represent the test partitions. In this sense, we ignore the comparisons between D14 and D16 because it does not make sense to the objective of this experiment given the composition of their train/test partitions.

\begin{table}[ht]
\centering
\renewcommand{\arraystretch}{1.4}
\begin{tabular}{lccccccccc}
\toprule[1pt]
\multirow{2}{*}{Training}	&		 & \multicolumn{7}{c}{Test} & \\ \cline{3-9}
        	 & & D14 & D16 & PL-I & PL-II & PHI & ES & SAM & Avg.\\ \hline
D14          & &  98.05	& --	& 51.09 & 50.51	& 89.45 & 76.28	& 81.50 & 74.48 \\
D16          & &   -- &	91.65 &	47.22 & 	46.59 &	87.04 &	76.10 &	81.73 & 71.72 \\
PL-I & & 32.90 &	15.40	&69.65&	69.33&	52.51	&0.23&	0.08 &	34.30 \\
PL-II & & 20.94	 & 10.52 & 69.46 & 69.13 & 50.23 & 0.54 & 0.24 & 31.58\\
PHI	  & & 88.75 & 72.17 & 42.12 & 41.89 & 93.82 & 73.20 & 77.42 & 69.91 \\
ES           & & 85.46 &	63.55	& 41.72 &	41.49 &	81.20	& 78.45 &	80.60 & 67.49 \\
SAM          & & 80.03 & 41.16 &	12.81 &	12.74 &	18.99 &	76.10 & 83.13 &	46.42 \\
\bottomrule[1pt]
\end{tabular}
\caption{Performance in terms of Fm achieved by an $\CAE$ trained with a specific training partition (rows) on the test partitions of the rest of datasets (columns).}
\label{tab:domain_adaptation}
\end{table}

These results give insights on the sensitivity of the model with respect to the similarity of the train and test document types. It can be observed that results are good when the $\CAE$ is trained with data of a similar domain. This is the case of results amongst text (D14, D16, and PHI), palm leaf (PL-I and PL-II), and musical scores (SAM and ES) documents. Likewise, this sensitivity is confirmed with the results obtained when models are trained with palm leaf documents (the most different to the rest), with which very poor results are achieved in the rest of the document types.

It is interesting to remark that the most general documents are the ones depicting Latin text (D14 and D16). Specifically, the model trained for D14 obtains a fair average performance considering all datasets, which is competitive against existing binarization algorithms (cf. Table \ref{tab:results_comparison}).

\section{Conclusions} \label{sec:conclusions}
In this paper an approach for document image binarization has been presented. The strategy is to train a Selectional Auto-Encoder ($\CAE$) that is able to learn an end-to-end transformation to binarize an image. Given a piece of image of a fixed size, the model outputs a selectional value for each pixel of the image depending on the confidence whether the pixel belongs to the foreground of the document. These values are eventually thresholded to yield a discrete binary result.

The influence of the model configuration and hyper-parameters has been studied preliminary. Then, the selected model has been evaluated under several datasets of distinct typology, as well as different ways of using the training set: using only domain-specific data or using all available data. The former option reported better results, yet the latter one also achieved a fair performance. 

The studied approach has been faced against both classical and state-of-the-art binarization algorithms. Results reported that the $\CAE$ approach is able to outperform the rest of binarization algorithms, increasing the best average performance obtained by existing strategies from $75.48$ up to $83.41$ in terms of F-measure.

The errors produced by the selected $\CAE$ have been analyzed in detail, drawing the conclusion that the approach detects the background very accurately and errors are concentrated around foreground strokes. Furthermore, it has been shown that errors are not directly related to their position within the input/output window.

Finally, a cross-document adaptation experiment has been carried out, in which we measured the performance of the models trained specifically for a dataset in the rest of them. The main conclusion drawn from this experiment is that the performance of the $\CAE$ is directly related to the similarity between train and test data. That is, good results are obtained if an appropriate training set is available for that domain.

As prospects of future work, we want to study how to include a priori information in the binarization model. For instance, it would be interesting to favor the continuity of ink strokes, so that it is more unlikely to find broken shapes in the foreground. In addition, in those cases when documents are too large to fit into a single window, it may be helpful to connect all the independent processes through recurrent connections so that information from all areas of the document are taken into account at once.

\section*{Acknowledgments}
This work was partially supported by the Social Sciences and Humanities Research Council of Canada, the Spanish Ministerio de Ciencia, Innovaci\'{o}n y Universidades through Juan de la Cierva - Formaci\'{o}n grant (Ref. FJCI-2016-27873), and the Universidad de Alicante through grant GRE-16-04.

\bibliographystyle{elsarticle-num}
\bibliography{binarization}

\end{document}